\pdfoutput=1

\documentclass[11pt]{article}

\usepackage{emnlp2021} %

\usepackage{times}
\usepackage{latexsym,float}

\usepackage[T1]{fontenc}

\usepackage[utf8]{inputenc}

\usepackage[english]{babel}

\usepackage{microtype}

\usepackage{todonotes}
\usepackage{tabularx}

\usepackage{booktabs,graphicx}
\usepackage{float}
\usepackage{enumitem}
\usepackage{array}

\usepackage{tablefootnote,booktabs}

\usepackage{todonotes,booktabs,csquotes,tabularx, multirow, graphicx,pgfplots, subfiles,tikzscale,pgfplots,footmisc}
\usepackage{dsfont,amsmath}

\usepackage{caption}
\usepackage{subcaption}

\usepackage[final]{listings}

\usepackage{array}
\newcolumntype{L}[1]{>{\raggedright\let\newline\\\arraybackslash\hspace{0pt}}m{#1}}
\newcolumntype{C}[1]{>{\centering\let\newline\\\arraybackslash\hspace{0pt}}m{#1}}
\newcolumntype{R}[1]{>{\raggedleft\let\newline\\\arraybackslash\hspace{0pt}}m{#1}}

\usepackage{svg}
\newcommand{\orcid}[1]{\href{https://orcid.org/#1}{\includegraphics[width=10pt]{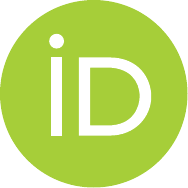}}}

\usepackage{hyperref}

\title{Towards Full-Fledged Argument Search: A 
Framework for Extracting and Clustering Arguments from Unstructured Text} %

\author{
    Michael F{\"a}rber\,\orcid{0000-0001-5458-8645} \and Anna Steyer \\
    Karlsruhe Institute of Technology (KIT), Karlsruhe, Germany \\
    \texttt{michael.faerber@kit.edu},\texttt{ annasteyer@outlook.com} \\
}

\begin{document}
\maketitle
\begin{abstract}
Argument search 
aims at identifying arguments in natural language texts. In the past, this task has been addressed by a combination of keyword search and argument identification on the sentence- or document-level. However, existing 
frameworks 
often address only specific components of argument search 
and 
do not address 
the following aspects:
(1) argument-query matching: identifying arguments that frame the topic slightly differently than the actual search query; 
(2) argument identification: identifying arguments that consist of multiple sentences; 
(3) argument clustering: selecting retrieved arguments by topical aspects. 
In this paper, we propose a framework for addressing these shortcomings. 
We suggest (1)~to combine the keyword search with precomputed topic clusters for argument-query matching, 
(2)~to apply a novel approach based on sentence-level sequence-labeling for argument identification, and 
(3)~to present aggregated arguments to users based on topic-aware argument clustering. 
Our 
experiments on several real-world debate data sets 
demonstrate that density-based clustering algorithms, such as {HDBSCAN}, 
are particularly suitable for 
argument-query matching. 
With our sentence-level, BiLSTM-based sequence-labeling approach 
we achieve a 
macro $F_1$ score of 0.71. 
Finally, evaluating our
argument clustering method 
indicates that a fine-grained clustering of arguments by subtopics 
remains challenging but is worthwhile to be explored. 
\end{abstract}

\section{Introduction}
\label{sec:Introduction}

Arguments are an integral part of 
debates and discourse between people. For instance, journalists, scientists, lawyers, and managers often need to pool arguments and contrast pros and cons \cite{DBLP:conf/icail/PalauM09a}. 
In light of this, argument search has been proposed to retrieve relevant arguments for a given query (e.g., \emph{gay marriage}).

Several argument search approaches have been proposed in the literature~\cite{Habernal2015,Peldszus2013}. 
However, major challenges of argument retrieval still exist, such as identifying and clustering arguments concerning controversial topics %
and extracting arguments from a wide range of texts on a fine-grained level. 
For instance, the argument search system args.me \cite{Wachsmuth2017} uses a keyword search to match relevant documents and lacks the extraction of individual arguments. In contrast, the ArgumenText \cite{Stab2018} applies keyword matching on single sentences to identify relevant arguments and neglects the context,  
yielding rather shallow arguments.
Furthermore, IBM Debater \cite{Levy2018} proposes a rule-based extraction of arguments from Wikipedia articles utilizing prevalent structures to identify sentence-level arguments.
Overall, the existing approaches are lacking (a)~a semantic argument-query matching, 
(b)~a segmentation of documents into argumentative units of arbitrary size, and (c)~a clustering of arguments 
w.r.t. subtopics 
for a fully equipped framework.

\begin{figure*}[tb]
\centering
\includegraphics[width=0.99\textwidth]{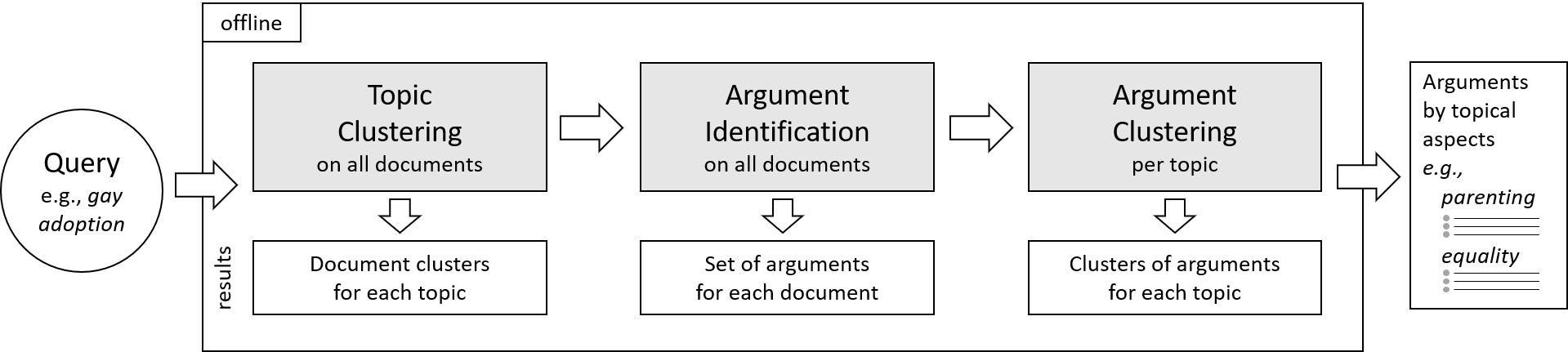}
\caption{Overview of our framework for argument search. %
}
\label{fig:Approach:Pipeline}
\end{figure*}

In this paper, we propose a novel argument search framework 
that addresses these aspects 
(see Figure~\ref{fig:Approach:Pipeline}): 
During the \textit{topic clustering} step, we group argumentative documents by topics and, thus, identify the set of %
relevant documents for a given search query 
(e.g., \textit{gay marriage}). To overcome the limitations of keyword search approaches, we rely on semantic representations via embeddings 
in combination with established clustering algorithms. 
Based on the relevant documents, the \textit{argument segmentation} step aims at identifying and separating arguments. We hereby understand arguments to consist of one or multiple sentences and propose a BiLSTM-based sequence labeling method. %
In the final \textit{argument clustering} step, we identify the different aspects of the topic at hand that are 
covered 
by the arguments identified in the previous step. 
We evaluate all three steps of our framework based on 
real-world data sets from several domains. 
By 
using the output of one framework's component as input for the subsequent component, 
our evaluation is 
particularly 
challenging but
realistic. In the evaluation, we show that by using embeddings, text labeling, and clustering, we can extract and aggregate arguments from unstructured text to a considerable extent. 
Overall, our framework 
provides the basis for advanced 
argument search in real-world scenarios with little training data.

In total, we make the following contributions:
\begin{itemize}
  \setlength\itemsep{0em}
\item We propose a novel argument search framework for fine-grained argument retrieval
based on topic clustering, argument identification, and argument clustering.\footnote{The source is available online at \url{https://github.com/michaelfaerber/arg-search-framework}.}
\item We provide a new evaluation data set for sequence labeling of arguments.
\item We evaluate all steps of our framework extensively based on four data sets. 

\end{itemize}

In the following, 
after discussing related work in Section~\ref{sec:RelatedWork}, 
we propose our argument mining framework in Section~\ref{sec:Approach}. 
Our evaluation is presented in Section~\ref{sec:Evaluation}. 
Section~\ref{sec:Conclusion} concludes the paper.

\section{Related Work}
\label{sec:RelatedWork}

\textbf{Topic Clustering. }
Various approaches for modeling the topics of documents have been proposed, such as Latent Dirichlet Allocation (LDA) and Latent Semantic Indexing (LSI). 
Topic detection and tracking~\cite{Wayne00} and topic segmentation~\cite{Ji2003} have been pursued in detail in the IR community.
\citet{Sun2007} introduce an unsupervised method for 
topic detection and topic segmentation of multiple similar documents. 
Among others, \citet{Barrow2020}, \citet{Arnold2019}, and \citet{Mota2019} propose models for segmenting documents and assigning topic labels to these segments, but ignore arguments.

\textbf{Argument Identification. }
\label{sec:RelatedWork:ArgumentRecognition}
Argument identification can be approached on the \textit{sentence level} by deciding for each sentence whether it constitutes an argument. For instance, IBMDebater \cite{Levy2018} relies on a combination of rules and weak supervision for classifying sentences as arguments. 
In contrast, ArgumenText \cite{Stab2018} does not limit its argument identification to sentences. 
\citet{Reimers2019} show that contextualized word embeddings can improve the identification of sentence-level arguments. 

Argument identification has been approached on the level of \textit{argument units}, too. Argument units are defined as different parts of an argument. 
\citet{Ajjour2017} compare machine learning techniques
for argument segmentation on several corpora. %
The authors observe that 
BiLSTMs mostly achieve the best results. %
Moreover, \citet{Eger2017} and \citet{Petasis2019} show 
that using more advanced models, such as combining a BiLSTM with CRFs and CNNs, hardly improves the BIO tagging results. Hence, we also create a BiLSTM model for argument identification.

\textbf{Argument Clustering.}
\label{sec:RelatedWork:ArgumentClustering}
\citet{Ajjour2019a} approach argument aggregation by identifying non-overlapping \emph{frames}, defined as a set of arguments from one or multiple topics that focus on the same aspect. 
\citet{Bar-Haim2020} propose an argument aggregation approach by mapping similar arguments to common \emph{key points}, i.e., high-level arguments. 
They observe that models with BERT embeddings perform the best for this task. 
\citet{Reimers2019} propose the clustering of arguments based on the similarity of two sentential arguments with respect to their topics. 
Also here, a fine-tuned 
BERT model is most successful for assessing the argument similarity automatically. %
Our framework, while being also based on BERT for argument clustering, can consist of several sentences, making it on the one hand more flexible but argument clustering on the other hand more challenging.

\textbf{Argument Search Demonstration Systems. } %
\label{sec:RelatedWork:Frameworks}
\citet{Wachsmuth2017} propose the argument search framework \textit{args.me}
using online debate platforms. %
The arguments are 
considered 
on document level. 
\citet{Stab2018} propose the framework \textit{ArgumenText} for argument search. %
The retrieval of topic-related web documents 
is based on keyword matching, while the argument identification is based on a binary sentence classification.
\citet{Levy2018} propose 
\textit{IBM Debater} based on Wikipedia articles. Arguments are defined as single claim sentences that explicitly discuss a \emph{main concept} in Wikipedia 
and that are identified via rules. 

\textbf{Argument Quality Determination.} 
Several approaches and data sets have been published on determining the quality of arguments  \cite{DBLP:conf/acl/GienappSHP20,DBLP:conf/cikm/DumaniS20,DBLP:conf/aaai/GretzFCTLAS20}, which is beyond the scope of this paper.  

\section{Approach}
\label{sec:Approach}
\label{sec:Approach:Overview}

In this section, we present the different steps of our framework: (1)~topic clustering, (2)~argument identification, and (3)~argument clustering according to topical aspects (see Figure~\ref{fig:Approach:Pipeline}).

\subsection{Topic Clustering}
\label{sec:Approach:DS}

First, documents are grouped into topics. Such documents can be individual texts or collections of texts under a common title, such as posts on Wikipedia or debate platforms. We compute the topic clusters using unsupervised clustering algorithms 
and study the results of \textit{k-means} and \textit{HDBSCAN}~\cite{Campello2013} in detail.
We also take the $argmax$ of the tf-idf vectors and LSA~\cite{Deerwester1990} vectors directly into consideration 
to evaluate how well topics are represented by single terms.

Overall, we consider the following models:

\begin{itemize}
\item $ARGMAX_{none}^{tfidf}$: We restrict the vocabulary size and the maximal document frequency to obtain a vocabulary representing topics with single terms. Thus, clusters are labeled with exactly one term by choosing the $argmax$ of these tf-idf document vectors.

\item $ARGMAX_{lsa}^{tfidf}$: We perform a dimensionality reduction with LSA on the tf-idf vectors. Therefore, each cluster is represented by a collection of terms.

\item $KMEANS_{none}^{tfidf}$: We apply the k-means clustering algorithm directly to tf-idf vectors and compare the results obtained by varying the parameter $k$.

\item $HDBSCAN_{umap}^{tfidf}$: We apply UMAP \cite{McInnes2018} dimensionality reduction on the tf-idf vectors. We then compute clusters using the HDBSCAN algorithm based on the resulting vectors.

\item $HDBSCAN_{lsa+umap}^{tfidf}$: Using the best parameter setting from the previous model, we apply UMAP dimensionality reduction on LSA vectors. Then, we evaluate the clustering results obtained with HDBSCAN while the number of dimensions of the LSA vectors is varied.

\end{itemize}

\subsection{Argument Identification}
\label{sec:Approach:SEG}

For the second step of our argument search framework, we propose the segmentation of sentences into argumentative units. 
Related works define arguments either on document-level~\cite{Wachsmuth2017} or sentence-level~\cite{Levy2018}\cite{Stab2018}, while, in this paper, we define an argumentative unit as a sequence of multiple sentences. This yields two advantages:
(1)~We can capture the context of arguments over multiple sentences (e.g., claim and its premises); 
(2)~Argument identification becomes applicable to a wide range of texts (e.g., user-generated texts).

Thus, we train a sequence-labeling model to predict for each sentence whether it 
starts an argument, continues an argument, or is outside of an argument (i.e., \textit{BIO-tagging}). Based on the findings of \citet{Ajjour2017}, \citet{Eger2017}, and \citet{Petasis2019}, we use a BiLSTM over more complex architectures like BiLSTM-CRFs. The BiLSTM is better suited than a LSTM as the bi-directionality takes both preceding and succeeding sentences into account for predicting the label of the current sentence. We evaluate the sequence labeling results compared to a feedforward neural network as baseline classification model that predicts the label of each sentence independently of the context.

We consider two ways to compute embeddings over sentences with BERT~\cite{Devlin2018}: 
\begin{itemize}
\item  \emph{bert-cls}, denoted as $MODEL_{cls}^{bert}$, uses the $[CLS]$ token corresponding output of BERT after processing a sentence.
\item \emph{bert-avg}, denoted as $MODEL_{avg}^{bert}$, uses the average of the word embeddings calculated with BERT as a sentence embedding. 
\end{itemize}

\subsection{Argument Clustering}
\label{sec:Approach:AS}

In the argument clustering task, we apply the same methods (k-means, DBSCAN) as in the topic clustering step to group the arguments within a specific topic by topical aspects. Specifically, we compute clusters of arguments for each topic and compare the performance of k-means and HDBSCAN with tf-idf, as well as \emph{bert-avg} and \emph{bert-cls} embeddings. Furthermore, we investigate whether calculating tf-idf within each topic separately is superior to computing tf-idf over 
all arguments in the document corpus (i.e., across topics).

\section{Evaluation}
\label{sec:Evaluation}

\subsection{Evaluation Data Sets}
\label{sec:Evaluation:Datasets}
In total, we use four data sets for evaluating the different steps of our argument retrieval framework.

\begin{enumerate}
\item \textbf{Debatepedia} is a debate platform that lists arguments to a topic on one
page, including subtitles, structuring the arguments into different aspects.\footnote{We use the data 
available at \url{https://webis.de/}.\label{dataset:webis}} 

 \item \textbf{Debate.org} is a debate platform that is organized in rounds where each of two opponents submits posts arguing for their side. Accordingly, the posts might also include non-argumentative parts used for answering the important points of the opponent before introducing new arguments.\footref{dataset:webis}
 
 \item \textbf{Student Essay} \cite{Stab2017} is widely used in research on argument segmentation \cite{Eger2017, Ajjour2017, Petasis2019}. Labeled on token-level, each document contains one major claim and several claims and premises. We can use this data set for evaluating argument identification.
 
 \item \textbf{Our Dataset} is based on a debate.org crawl.\footref{dataset:webis}  It is restricted to a subset of four out of the total 23 categories -- \textit{politics}, \textit{society}, \textit{economics} and \textit{science} -- and contains additional annotations. 
 3 human annotators familiar with linguistics segmented these documents 
and labeled them as being of \emph{medium} or \emph{low quality}, to exclude low quality documents. 
The annotators were then asked to indicate the beginning of each new argument and to label argumentative sentences summarizing the aspects of the post as \emph{conclusion} and \emph{outside of argumentation}. In this way, we obtained a ground truth of labeled arguments on a sentence level (Krippendorff's $\alpha=0.24$ based on 20 documents and three annotators). 
A 
description of the data set is provided online.\footnote{See 
\url{https://github.com/michaelfaerber/arg-search-framework}.}
\end{enumerate}

\textbf{Train-Validation-Test Split.} We use the splits provided by \citet{Stab2017} for the student essay data set. The other data sets are divided into train, validation and test splits based on topics (15\% of the topics were used for testing).

\begin{table*}[tb]
\centering
\caption{Results of unsupervised topic clustering on the \emph{debatepedia} data set $-$ \%$n$:~noise examples (HDBSCAN)
}
\label{tab:Results:ClusteringUnsupervised}
\begin{small}
\resizebox{0.99\textwidth}{!}{
\begin{tabular}{@{}l rrrrr  r rrrrr@{}}
\toprule
& \multicolumn{5}{c}{With noise}& &\multicolumn{5}{c}{\emph{Without noise}}\\
\cline{2-6} \cline{8-12}
& \#$Clust.$ & $ARI$ &$Ho$ & $Co$ & $BCubed~F_1$ & \%$n$ & \#$Clust.$ & $ARI$ &$Ho$ & $Co$ & $F_{1}$ \\
\midrule
$ARGMAX_{none}^{tfidf}$
&253 	& 0.470 &0.849 & 0.829	& 0.591 & $-$ & $-$ & $-$ & $-$ & $-$ & $-$\\

$ARGMAX_{lsa}^{tfidf}$
&157	& 0.368 & 0.776& 0.866	& 0.561& $-$ & $-$ & $-$ & $-$ & $-$ & $-$\\

$KMEANS_{none}^{tfidf}$
&170 	& \textbf{0.703} & 0.916 &	0.922	& 0.774 & $-$ & $-$ & $-$ & $-$ & $-$ & $-$ \\

$HDBSCAN_{none}^{tfidf}$
& 206 & 0.141  &0.790 & 0.870	& 0.677 & 21.1 &205 & 0.815 &0.955& 0.937	& 0.839\\

$HDBSCAN_{umap}^{tfidf}$
& 155 & 0.673  &0.900 &	0.931	& 0.786 & 4.3 & 154 & 0.779 &0.927&  0.952& 0.827\\

$HDBSCAN_{lsa + umap}^{tfidf}$
&162 	& 0.694 & 0.912& 0.935& \textbf{0.799} & 3.6 &161 & 0.775 &0.932 &0.950	& 0.831\\
\bottomrule
\end{tabular}
}
\end{small}
\end{table*}

\begin{table*}[tb]
\centering
\caption{Results of unsupervised topic clustering on the \emph{debate.org} data set
$-$ \%$n$:~noise examples (HDBSCAN)}
\label{tab:Results:ClusteringUnsupervisedORG}
\begin{small}
\resizebox{0.99\textwidth}{!}{
\begin{tabular}{@{}l  rrrrr  r rrrrr@{}}
\toprule
& \multicolumn{5}{c}{With noise}& &\multicolumn{5}{c}{\emph{Without noise}}\\
\cline{2-6} \cline{8-12}
& \#$Clust.$ & $ARI$  & $Ho$ &$Co$& $F_{1}$ & \%$n$ &\#$Clust.$ & $ARI$ & $Ho$ &$Co$ &  $BCubed~F_1$ \\
\midrule

$KMEANS_{none}^{tfidf}$ & 50 & \textbf{0.436} & 0.822 & 0.796 & \textbf{0.644} &$-$&$-$&$-$&$-$&$-$&$-$\\

$HDBSCAN_{umap}^{tfidf}$
& 20 & 0.354 &0.633& 0.791 & 0.479 & 7.1 & 19 & 0.401 & 0.648 & 0.831 & 0.502\\

$HDBSCAN_{lsa + umap}^{tfidf}$
&26 	& 0.330 & 0.689 &0.777 & 0.520 & 5.8 & 25 & 0.355 & 0.701 & 0.790 &0.542\\

\bottomrule
\end{tabular}
}
\end{small}
\end{table*}

\subsection{Evaluation Settings}
\label{sec:Evaluation:Methods}

We report the results by using the evaluation metrics 
precision, recall and $F_1$-measure concerning the classification, and adjusted rand index (ARI), homogeneity (Ho), completeness (Co), and $BCubed~F_1$ score concerning the clustering.

\textbf{Topic Clustering.} We use HDBSCAN with the minimal cluster size of 2. 
Regarding k-means, 
we vary the parameter $k$ that determines the number of clusters. We only report the results of the best setting.
For the model $ARGMAX_{none}^{tfidf}$ we restrict the vocabulary size and the maximal document frequency of the tf-idf to obtain a vocabulary that best represents the topics by single terms. %

\textbf{Argument Identification.}
We use 
a BiLSTM implementation with 200 hidden units and apply \textit{SigmoidFocalCrossEntropy} 
as loss function. 
Furthermore, we use the \emph{Adagrad} optimizer~\cite{Duchi2011} and train the model for 600 epochs, shuffling the data in each epoch, and keeping only the best-performing model as assessed on the validation loss. The baseline feed-forward neural network contains a single hidden dense layer of size 200 and gets compiled with the same hyperparameters. 
As BERT implementation, we use DistilBERT~\cite{Sanh2019}. %

\begin{figure}[tb]
\centering
\includegraphics[width=0.78\linewidth]{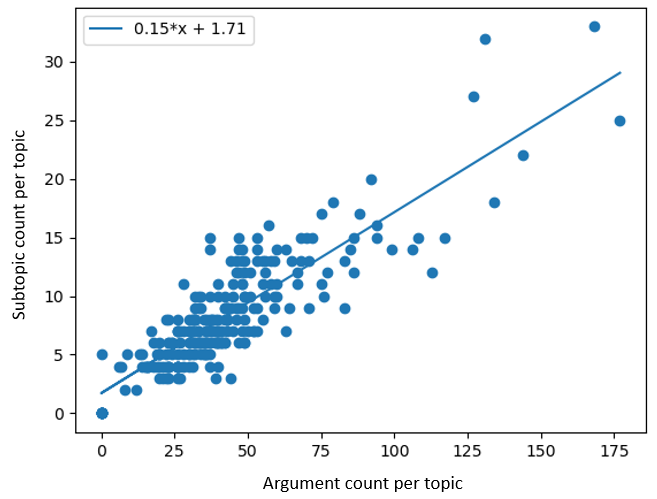}
\caption{Linear regression of arguments and their topical aspects for the \emph{debatepedia} data set.}%
\label{fig:Results:Correlation}
\end{figure}

\textbf{Argument Clustering.}
We estimate the parameter $k$ for the k-means algorithm for each topic using a linear regression based on the number of clusters relative to the number of arguments in this topic. As shown in Figure \ref{fig:Results:Correlation}, we observe a linear relationship between the number of topical aspects (i.e., subtopics) and the argument count per topic in the \emph{debatepedia} data set. 
We apply HDBSCAN with the same parameters as in the topic clustering task (\emph{min\_cluster\_size}~=~2).

\subsection{Evaluation Results}
\label{sec:Results}

In the following, we present the results for the three tasks topic clustering, argument identification, and argument clustering.

\subsubsection{Topic Clustering}
\label{sec:Results:DS}
We start with the results of clustering the documents by topics based on 
the \emph{debatepedia} data set with 170 topics (see Table \ref{tab:Results:ClusteringUnsupervised}). The plain model $ARGMAX_{none}^{tfidf}$, which computes clusters based on words with the highest tf-idf score within a document, achieves an $ARI$ score of $0.470$. This indicates that many topics in the \emph{debatepedia} data set can already be represented by a single word. Considering the $ARGMAX_{lsa}^{tfidf}$ model, the $ARI$ score of $0.368$ shows that using the topics found by LSA does not add value compared to tf-idf. 

Furthermore, we 
find that the tf-idf-based 
$HDBSCAN_{lsa + umap}^{tfidf}$ and $KMEANS_{none}^{tfidf}$ achieve a comparable performance given the $ARI$ scores of $0.694$ and $0.703$. However, since HDBSCAN accounts for noise in the data which it pools within a single cluster, the five rightmost columns of Table \ref{tab:Results:ClusteringUnsupervised} need to be considered when deciding on a clustering method for further use. When excluding the noise cluster from the evaluation, 
the $ARI$ score increases considerably for all $HDBSCAN$ models ($HDBSCAN_{none}^{tfidf}$: $0.815$, $HDBSCAN_{umap}^{tfidf}$: $0.779$, $HDBSCAN_{lsa+umap}^{tfidf}$: $0.775$). Considering that $HDBSCAN_{none}^{tfidf}$ identifies 21.1\% of the instances as noise, compared to the 4.3\% 
in case of $HDBSCAN_{umap}^{tfidf}$, we conclude that applying an UMAP dimensionality reduction step before the HDBSCAN clustering is quite fruitful for the performance (at least for the \emph{debatepedia} data set).

\begin{table*}[tb]
\centering
\caption{Results of argument identification on the \emph{Student Essay} data set.
}
\label{tab:Results:Segmentation}
\begin{small}
\begin{tabular}{@{}l cc cc cc ccc@{}}
\toprule
& \multicolumn{2}{c}{\emph{B}}& \multicolumn{2}{c}{\emph{I}}& \multicolumn{2}{c}{\emph{O}}& \multicolumn{2}{r}{}\\ \cline{2-3} \cline{4-5} \cline{6-7}

 & $Prec.$ & $Rec.$  
 & $Prec.$ & $Rec.$  
 & $Prec.$ & $Rec.$  
 &$F_{1, macro}$ & $F_{1, weighted}$ &$F_{1, macro}^{B, I}$ \\
\midrule
majority class 
& 0.000 & 0.000
& 0.719 & 1.000 
& 0.000 & 0.000 
&	0.279 & 0.602 & 0.419\\
$FNN_{avg}^{bert}$  
& 0.535 & 0.513 
& 0.820 & 0.836	
& 0.200 & 0.160 
& 0.510 & 0.736 & 0.675\\
$FNN_{cls}^{bert}$  
& 0.705 & 0.593	
& 0.849 & 0.916	
& \textbf{0.400} & 0.080 
& 0.553 & 0.805 & 0.763\\
$BILSTM_{avg}^{bert}$  
& 0.766 & 0.713 
& 0.885 & 0.930 
& 0.000 & 0.000 
& 0.549 & 0.846 & 0.823 \\
$BILSTM_{cls}^{bert}$  
 & \textbf{0.959}  & \textbf{0.914}
 & \textbf{0.967}  & \textbf{0.985}
 & 0.208 & \textbf{0.200}
& \textbf{0.705} & \textbf{ 0.951} & \textbf{0.956} \\
\bottomrule
\end{tabular}
\end{small}
\end{table*}

\begin{table*}[tb]
\centering 
\caption{Argument identification results on the \emph{debate.org} data set; model trained on \emph{Student Essay} and \emph{debate.org}.}
\label{tab:Results:SegORG}
\begin{small}
\begin{tabular}{@{}ll cc cc cc cc@{}}
\toprule
& & \multicolumn{2}{c}{\emph{B}}& \multicolumn{2}{c}{\emph{I}}& \multicolumn{2}{c}{\emph{O}}& \multicolumn{2}{r}{}\\ 
\cline{3-8}
&\emph{trained on}
 & $Prec.$ & $Rec.$  
 & $Prec.$ & $Rec.$  
 & $Prec.$ & $Rec.$  
 &$F_{1, macro}$ & $F_{1, weighted}$ \\
\midrule
$BILSTM_{cls}^{bert}$ & \emph{Student Essay}
& 0.192 & 0.312 
& 0.640 & 0.904 
& 1.000 & 0.015 
& 0.339 & 0.468\\
$BILSTM_{cls}^{bert}$  &\emph{debate.org}
& 1.000 & 0.021 
& 0.646 & 0.540 
& 0.381 & 0.640 
& 0.368 & 0.492\\
\bottomrule
\end{tabular}
\end{small}
\end{table*}

\textbf{Inductive Generalization.}
We apply our unsupervised approaches to the \emph{debate.org} data set to evaluate whether the model is able to generalize. The results, given in Table \ref{tab:Results:ClusteringUnsupervisedORG}, show that 
k-means performs distinctly better on the \emph{debate.org} data set than HDBSCAN (ARI: $0.436$ vs. $0.354$ and $0.330$; $F_1$: $0.644$ vs. $0.479$ and $0.520$). This is likely because the \emph{debate.org} data set is characterized by a high number of single-element topics. In contrast to k-means, HDBSCAN does not allow single-element clusters, thus getting pooled into a single noise cluster. This is reflected by the different numbers of clusters as well as the lower homogeneity scores of HDBSCAN of 0.633 and 0.689 compared to 0.822 of k-means while having a comparable completeness of approx. 0.8.

\textbf{Qualitative Evaluation.} 
Applying a keyword search would retrieve only a subset of the relevant arguments that mention the search phrase explicitly. Combining a keyword search on documents with the computed clusters 
enables an argument search framework to retrieve arguments from a broader set of documents. For example, 
for \textit{debatepedia.org}, we observe that clusters of arguments related to \emph{gay adoption} include words like \emph{parents}, \emph{mother}, \emph{father},  \emph{sexuality}, \emph{heterosexual}, and \emph{homosexual} while neither \emph{gay} nor \emph{adoption} are mentioned explicitly.

\subsubsection{Argument Identification}
\label{sec:Results:SEG}

The results of the argument identification step based on the \emph{Student Essay} data set are given in Table~\ref{tab:Results:Segmentation}. We evaluate a BiLSTM model in a sequence-labeling setting compared to 
majority voting and 
a feedforward neural network (FNN) as baselines. 
Using the BiLSTM improves the macro $F_1$ score relative to the feedforward neural network on the \emph{bert-avg} embeddings by 3.9\% and on the \emph{bert-cls} embeddings by 15.2\%. Furthermore, 
using \emph{bert-cls} embeddings increases the macro $F_1$ score by 4.3\% in the classification setting and by 15.6\% in the sequence-learning setting compared to using \emph{bert-avg}. 

\textbf{BIO Tagging.}
We observe a low precision and recall for the class `O'. 
This can be traced back to a peculiarity of the \emph{Student Essay} data set: Most sentences in the \emph{Student Essay}'s training data set are part of an argument and only 3\% of the sentences are labeled as non-argumentative (outside/`O'). Accordingly, the models observe hardly any examples for outside sentences during training and, thus, have difficulties in learning to distinguish them from other sentences. 
Considering the fact that the correct identification of a `B' sentence alone is already enough to separate two arguments from each other, the purpose of labeling a sentence as `O' is restricted to classifying the respective sentence as non-argumentative. Therefore, in case of the \emph{Student Essay} data set, the task of separating two arguments from each other becomes much more important than separating non-argumentative sentences from arguments. In the last column of Table \ref{tab:Results:Segmentation}, we also show the macro $F_1$ score for the `B' and `I' labels only. The high macro $F_1$ score of 0.956 for the best performing model reflects the model's high ability to separate arguments from each other.

\begin{table}
\centering
\caption{Confusion matrices of $BILSTM_{cls}^{bert}$ (rows: ground-truth labels; columns: predictions).}
\label{tab:Results:CM}
\begin{footnotesize}
\begin{tabular}{c| ccc}
\multicolumn{4}{c}{Student Essay} \\
& B & I & O\\
\midrule
B&15 	& 33 	& 0\\
I&24	& 226	& 0 \\
O&39 	& 94 	& 2\\
\bottomrule
\end{tabular}
\hspace{1cm}
\begin{tabular}{c| ccc}
\multicolumn{4}{c}{debate.org} \\
& B & I & O\\
\midrule
B&1 	& 24 & 23\\
I&0	& 135 & 115 \\
O&0 	& 50 	& 85\\
\bottomrule
\end{tabular}
\end{footnotesize}
\end{table}

\begin{table*}[tb]
\centering
\caption{Results of the clustering of arguments of the \emph{debatepedia} data set by topical aspects. \emph{across topics}:~tf-idf scores are computed across topics, \emph{without noise}:~{HDBSCAN} is only evaluated on instances not classified as noise.}
\label{tab:Results:ASUnsupervised}
\begin{small}
\begin{tabular}{@{}l l l cccc l@{}}
\toprule
 Embedding & Algorithm & Dim. Reduction & $ARI$ & $Ho$	&$Co$	& $BCubed~F_1$ & Remark\\
\midrule
tf-idf 		& HDBSCAN 	& UMAP 	& 0.076 & 0.343	&0.366& 0.390 & \\
tf-idf 		& HDBSCAN 	& UMAP 	& 0.015 & 0.285	&0.300& 0.341 &\emph{across topics} \\
tf-idf 		& HDBSCAN 	& $-$ 	& \textbf{0.085} & \textbf{0.371}	&\textbf{0.409}& \textbf{0.407}	& \\
tf-idf 		& k-means 	& $-$ 	& 0.058 & 0.335	&0.362& 0.397 & \\
tf-idf 		& k-means 	& $-$ 	& 0.049 & 0.314	&0.352& 0.402 &\emph{across topics} \\
\midrule
bert-cls 	& HDBSCAN 	& UMAP 	& 0.030	& 0.280	&0.298& 0.357 & \\
bert-cls 	& HDBSCAN 	& $-$		& 0.016	& 0.201	&0.324& 0.378 & \\
bert-cls 	& k-means 	& $-$		& 0.044	& 0.332	&0.326& 0.369 & \\
\midrule
bert-avg 	& HDBSCAN 	& UMAP 	& 0.069	& 0.321	&0.352& 0.389 & \\
bert-avg 	& HDBSCAN 	& $-$		& 0.018	& 0.170	&0.325& 0.381 & \\
bert-avg 	& k-means 	& $-$		& 0.065	& 0.337	&0.349& 0.399 & \\
\midrule
tf-idf 		& HDBSCAN 	& $-$ 	& 0.140 & 0.429	&0.451& 0.439	&\emph{without noise}\\
\bottomrule
\end{tabular}
\end{small}
\end{table*}

\textbf{Generalizability.} 
We evaluate whether the 
model $BILSTM_{cls}^{bert}$, which performed best on the \emph{Student Essay} data set, is able to identify arguments on the \emph{debate.org} data set if trained on the \emph{Student Essay} data set. The results are given in Table~\ref{tab:Results:SegORG}.
Again, the pretrained model performs poor on \emph{`O'} sentences since not many examples of \emph{`O'} sentences were observed during training. Moreover, applying the pretrained $BILSTM_{cls}^{bert}$ to the \emph{debate.org} data set yields low precision and recall on \emph{`B'} sentences. A likely reason is that, in contrast to the \emph{Student Essay} data set where arguments often begin with cue words (e.g., \emph{first}, \emph{second}, \emph{however}), the documents in the \emph{debate.org} data set contain cue words less often. %

The results from training the $BILSTM_{cls}^{bert}$ model from scratch on the \emph{debate.org} data set are considerably different to our results on the \emph{Student Essay} data set. As shown in Table \ref{tab:Results:CM}, the confusion matrix for the \emph{debate.org} data set shows that the BiLSTM model has difficulties to learn which sentences start an argument in the \emph{debate.org} data set. In contrast to the \emph{Student Essay} data set, it cannot learn from the peculiarities (e.g., cue words) and apparently fails to find other indications for \emph{`B'} sentences. In addition, also the distinction between \emph{`I'} and \emph{`O'} sentences is not clear. These results match our experiences with the annotation of documents in the \emph{debate.org} data set where it was often difficult to decide whether a sentence forms an argument and to which argument it belongs. This is also reflected by the inter-annotator agreement of 0.24 based on Krippendorff's $\alpha$ on a subset of 20 documents with three annotators.

\textbf{Bottom Line.} Overall, we find that the performance of the argument identification strongly depends on the peculiarities and quality of the underlying data set. For well curated data sets such as the \emph{Student Essay} data set, the information contained in the current sentence as well as the surrounding sentences yield a considerably accurate identification of arguments. In contrast, data sets with poor structure or colloquial language, as given in the \emph{debate.org} data set, lead to less accurate results.%

\subsubsection{Argument Clustering}
\label{sec:Results:AS}

We now evaluate the argument clustering according to topical aspects (i.e., subtopics) as the final step of the argument search framework, 
using the \emph{debatepedia} data set. 
We evaluate the performance of the clustering algorithms HDBSCAN and k-means for different 
embeddings that yielded the best results in the topic clustering step of our framework. 
We perform the clustering of the arguments for each 
topic (e.g., \textit{gay marriage}) separately and average the results across the topics. As shown in Table \ref{tab:Results:ASUnsupervised}, we observe that HDBSCAN performs best on tf-idf embeddings with an averaged $ARI$ score of 0.085 while k-means achieves its best performance on \emph{bert-avg} embeddings with an averaged $ARI$ score of 0.065. Using HDBSCAN instead of k-means on tf-idf embeddings yields an improvement in the $ARI$ score of 2.7\%. Using k-means instead of HDBSCAN on \emph{bert-avg} and  \emph{bert-cls} embeddings results in slight improvements. %

\textbf{UMAP.} Using an UMAP dimensionality reduction step before applying HDBSCAN outperforms k-means on \emph{bert-avg} embeddings with an $ARI$ score of 0.069. However, using a UMAP dimensionality reduction in combination with tf-idf results in a slightly reduced performance.
We 
find that \emph{bert-avg} embeddings result in slightly better scores than \emph{bert-cls} when using UMAP. 

\textbf{TF-IDF across Topics.} We further evaluate whether computing tf-idf within each topic separately leads to a better performance than computing tf-idf  
across topics in the data set. 
The observed slight deviation of the $ARI$ score 
for k-means and 
HDBSCAN in combination with UMAP matches our expectation that 
clustering algorithms focus more on terms which distinguish the arguments from each other within a topic. 

\textbf{Excluding Noise.} When excluding the HDBSCAN noise clusters (\emph{without noise}), we yield an $ARI$ score of 0.140 and a $BCubed~F_1$ score of 0.439.

\begin{figure}[tb]
\centering
\includegraphics[width = 0.83\linewidth]{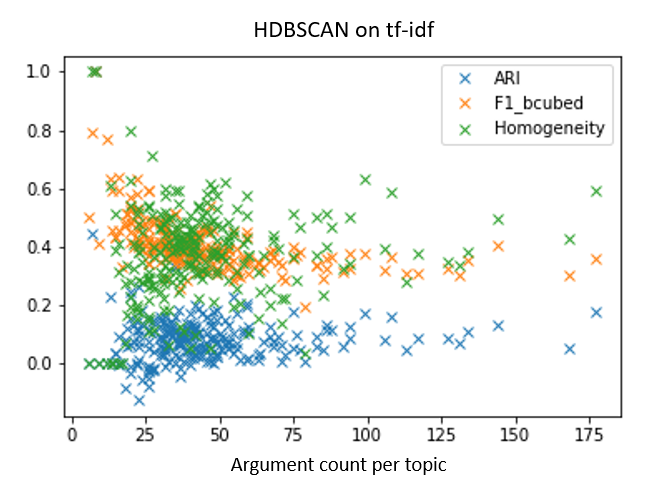}
\includegraphics[width = 0.83\linewidth]{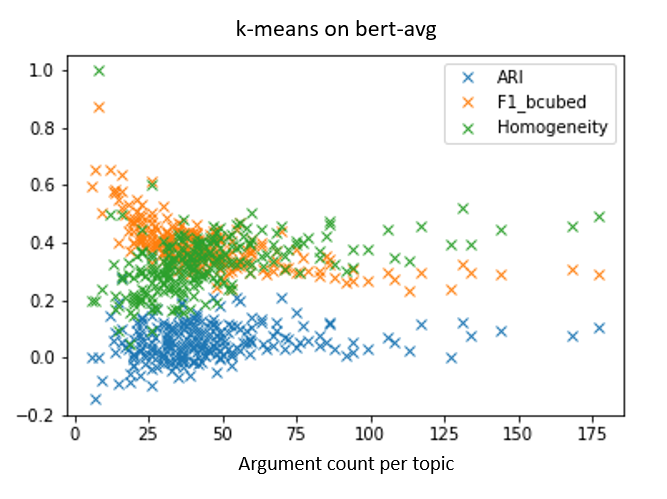}
\caption{Number of arguments in each topic
for HDBSCAN with tf-idf embeddings and k-means with \emph{bert-avg} embeddings.}
\label{fig:Results:Scatterplot}
\end{figure}

\textbf{Number of Arguments.} Figure~\ref{fig:Results:Scatterplot} shows the performance of the models HDBSCAN with tf-idf and k-means with \textit{bert-avg} with respect to the number of arguments in each topic. Both $ARI$ and $BCubed~F_1$ scores show a similar distribution for topics with different numbers of arguments, while the distributions of the homogeneity score show a slight difference for the two models. This indicates that the performance of the clustering algorithms does not depend on the number of arguments. 

\textbf{Examples.} In Table \ref{tab:Results:BestTopics}, we show for the best performing k-means and HDBSCAN models the topics with the highest $ARI$ scores.

\begin{table}[tb]
\centering
\caption{Top 5 topics 
using HDBSCAN and k-means.} %
\label{tab:Results:BestTopics}
\begin{small}
\begin{tabular}{p{3.9cm} r@{}r}
\toprule
Topic & \#$Clust._{true}$ & \#$Clust._{pred}$ \\
\toprule
\multicolumn{3}{c}{\textbf{HDBSCAN based on tf-idf embeddings}}\\
\midrule
Rehabilitation vs retribution
&2 &3  \\
Manned mission to Mars
&5 &6 \\
New START Treaty
&5 &4  \\
Obama executive order to raise the debt ceiling
&3 &6  \\
Republika Srpska secession from Bosnia and Herzegovina
&4 &6  \\
\toprule
\multicolumn{3}{c}{\textbf{k-means based on \emph{bert-avg} embeddings}}\\
\midrule

Bush economic stimulus plan
&7 &5  \\
Hydroelectric dams
&11 &10 \\
Full-body scanners at airports
&5 &6 \\
Gene patents
&4 &7  \\
Israeli settlements
&3 &4 \\
\bottomrule
\end{tabular}
\end{small}
\end{table}

\textbf{Bottom Line.} 
Overall, our results confirm that argument clustering based on topical aspects is nontrivival and high evaluation results are still hard to achieve in real-world settings.  
Given the \emph{debatepedia} data set, we show that our unsupervised clustering algorithms with the different embedding methods do not cluster arguments into topical aspects in a highly consistent and reasonable way yet. This result is in line with the results of \citet{Reimers2019} stating that even experts have difficulties to identify argument similarity based on topical aspects (i.e., subtopics).
Considering that their evaluation is based on sentence-level arguments, it seems likely that assessing argument similarity is even harder for arguments comprised of one or multiple sentences.
Moreover, the authors report promising results for the pairwise assessment of argument similarity when using the output corresponding to the BERT $[CLS]$ token. 
However, our experiments show that their findings do not apply to the \emph{debatepedia} data set. We assume that this is due to differences in the argument similarity that are introduced by using prevalent topics in the \emph{debatepedia} data set rather than using explicitly annotated arguments.

\section{Conclusion} %
\label{sec:Conclusion}

In this paper, we proposed an   
argument search framework that 
combines the keyword search with precomputed topic clusters for argument-query matching, applies a novel approach to argument identification based on sentence-level sequence labeling, 
and 
aggregates arguments via argument clustering. 
Our evaluation with real-world data 
showed that our framework can be used 
to mine and search for arguments from unstructured text 
on any given topic. 
It 
became clear that a full-fledged argument search requires a deep understanding of text and that the individual steps can still be improved. 
We suggest future research on developing argument search approaches that are sensitive to different aspects of argument similarity and argument quality. %

\bibliography{bibliography}
\bibliographystyle{acl_natbib}

\end{document}